\documentclass[letterpaper, 10 pt, conference]{ieeeconf}  

\IEEEoverridecommandlockouts                              

\usepackage{graphicx}
\graphicspath{{figures/}}

\usepackage{amsmath} 
\usepackage{amssymb}  
\usepackage{balance}
\usepackage{pgfplots}
\usepackage{subcaption}

\usepackage[separate-uncertainty = true]{siunitx}
\usepackage[hidelinks]{hyperref}
\usepackage[capitalise]{cleveref}
\Crefformat{figure}{#2Fig.~#1#3}
\Crefmultiformat{figure}{Figs.~#2#1#3}{ and~#2#1#3}{, #2#1#3}{ and~#2#1#3}

\usepackage[textsize=tiny]{todonotes}

\newcommand{\mm}[1]{\textcolor{black}{#1}}

\title{\LARGE \bf
Planar Modeling and Sim-to-Real of a Tethered Multimaterial Soft Swimmer Driven by Peano-HASELs
}

\author{Stephan-Daniel Gravert$^{*}$, Mike Y. Michelis$^{*}$, Simon Rogler, Dario Tscholl, Thomas Buchner,\\ and Robert K. Katzschmann
\thanks{*These authors contributed equally to this work.}
\thanks{All authors are with the Soft Robotics Lab, Department of Mechanical and Process Engineering,
        ETH Zurich, Tannenstrasse 3, 8092 Zurich, Switzerland
        \tt \footnotesize 
\{\href{mailto:sgravert@ethz.ch}{sgravert},
\href{mailto:mmichelis@ethz.ch}{michelism},
\href{mailto:srogler@ethz.ch}{srogler},
\href{mailto:dtscholl@ethz.ch}{dtscholl},
\href{mailto:tbuchner@ethz.ch}{tbuchner},
\href{mailto:rkk@ethz.ch}{rkk}\}@ethz.ch}
}

\begin{document}

\maketitle
\thispagestyle{empty}
\pagestyle{empty}

\begin{abstract}
Soft robotics has the potential to revolutionize robotic locomotion, in particular, soft robotic swimmers offer a minimally invasive and adaptive solution to explore and preserve our oceans. Unfortunately, current soft robotic swimmers are vastly inferior to evolved biological swimmers, especially in terms of controllability, efficiency, maneuverability, and longevity. Additionally, the tedious iterative fabrication and empirical testing required to design soft robots has hindered their optimization. In this work, we tackle this challenge by providing an efficient and straightforward pipeline for designing and fabricating soft robotic swimmers equipped with electrostatic actuation. We streamline the process to allow for rapid additive manufacturing, and show how a differentiable simulation can be used to match a simplified model to the real deformation of a robotic swimmer. We perform several experiments with the fabricated swimmer by varying the voltage and actuation frequency of the swimmer's antagonistic muscles. We show how the voltage and frequency vary the locomotion speed of the swimmer while moving in liquid oil and observe a clear optimum in forward swimming speed. 
The differentiable simulation model we propose has various downstream applications, such as control and shape optimization of the swimmer; optimization results can be directly mapped back to the real robot through our sim-to-real matching.

\end{abstract}


\section{INTRODUCTION}


\subsection{Motivation} 
Robots truly useful to humans must integrate into our world by safely performing tasks in close proximity to living beings. Ideally, robots seamlessly integrate into our daily lives and assist us without any possibility of causing physical harm. While rigid robots are powerful and fast~\cite{kuindersma_optimization-based_2016, miki2022learning}, they lack dexterity and cannot adapt well to their environment; thus, they can be dangerous to humans. In contrast, soft robots~\cite{rus_design_2015} have a partially compliant and multifunctional composition and the potential to self-heal, biodegrade, and interact safely with their environment. Therefore, soft robots will likely be the only robots to fully integrate and perform smoothly in the real world~\cite{hawkes_hard_2021}. Soft robots are inspired by animals’ ability to deform, adapt, and survive in complex environments~\cite{morin_camouflage_2012, miriyev_soft_2017}. 

Indeed, technologies enabling robotic movement based on pneumatic, magnetic, thermal, and electrostatic actuation mechanisms~\cite{miriyev_soft_2017, sachyani_keneth_multi-material_2020, ilami_materials_2020} have been investigated with the aim of creating soft robots with the ability to move, adapt, and interact with complex and unpredictable natural environments in the same way as animals~\cite{berlinger_fish-like_2021, costa_design_2020, li_self-powered_2021, scaradozzi_bcf_2017}. The replication of animals’ movement abilities in robots would revolutionize robotic locomotion, and bring us closer to safe robot companions. However, no computational tool to date can model and optimize soft robot performance. 

\begin{figure}[t]%
    \centering
    \includegraphics[width=\columnwidth]{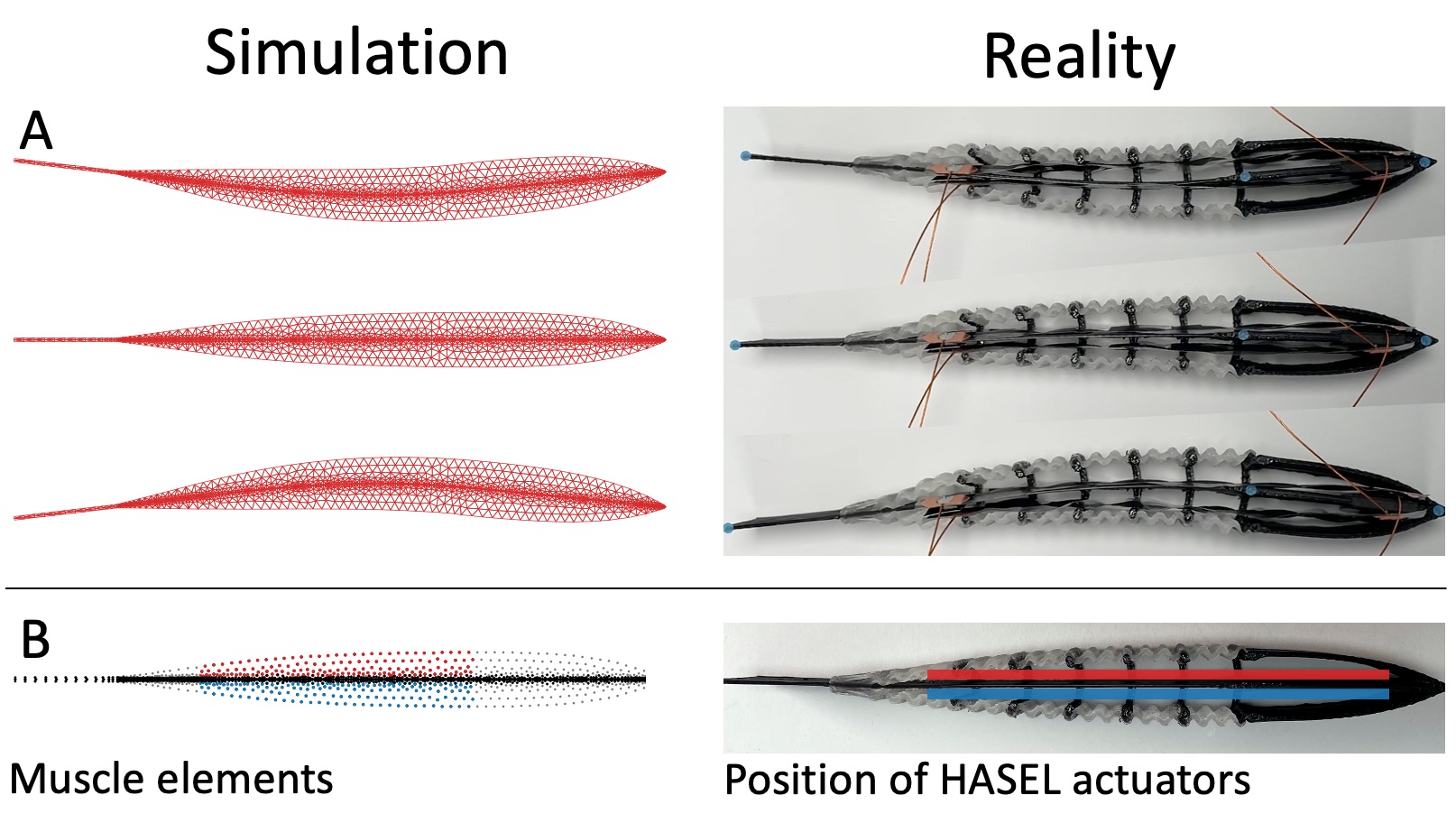}
    \caption{(A) Comparison of our electrostatically-actuated swimmer with simulation, actuated at 5.5kV in both simulation and reality. (B) Comparison of actuator position in simulation versus reality. In simulation we depict the vertices, which are colored according to which type of element they belong to: red is upper muscle, blue is lower muscle, black is spine elements, and grey is soft material in the swimmer body.}
    \label{fig:simvsreal}
\end{figure}

Soft robots have attracted particular attention for their potential applications within fluidic environments. For instance, soft robots moving within aquatic environments can serve for naturalistic sampling, geographical mapping, and exploration of oceans both on earth and in outer space~\cite{berlinger_fish-like_2021, costa_design_2020, li_self-powered_2021, scaradozzi_bcf_2017}. Oceans are filled with objects and living entities still to be discovered as only 20\% of it been mapped to date~\cite{trethewey_earths_2020}. Soft robotic swimmers offer a minimally invasive, biodegradable and affordable solution for exploration of this ‘final frontier’ on earth. Biomimetic soft robots could integrate into an ecosystem to explore, observe, and preserve biodiversity~\cite{krause_interactive_2011}, natural resources~\cite{marx_sargassum_2021}, and new habitats~\cite{gomez_jr_size_2017}. Soft robots could potentially observe vulnerable populations in protected areas where only minimally-invasive biomimetic~\cite{marras_fish_2012, butail_fishrobot_2015} and biodegradable~\cite{aracri_soft_2021} entities are permitted. To therefore develop efficient soft robots for these challenges, we need effective strategies that can quickly prototype and optimize robots for their target objective.

\subsection{Approach}
We propose an electrostatically-actuated soft robotic swimmer using Hydraulically amplified self-healing electrostatic (HASEL) actuators, manufactured using rapid multimaterial 3D-printing technology, and modeled in a differentiable simulation using simplified muscle models that close the sim-to-real gap through gradient-based optimization. HASELs are a promising soft actuator technology due to their muscle-like performance. Combined with soft-rigid printing for a robot ``body'', we can create soft robots in a swift manner. Since it is often tedious to optimize for control and shape of the resulting robot, we integrate our model into a differentiable simulation with simulated muscle models that match the real actuation behavior such that it can be used in future downstream gradient-based optimization tasks.

\subsection{Contributions}

We developed (1) a 2.5D (2D shape extruded with constant height profile) Peano HASEL soft robotics swimmer using (2) a rapid multimaterial manufacturing approach for soft robotic systems, and \mm{(3) present a Finite Element Method (FEM) model for HASEL actuators that matches a simplified muscle model in a differentiable simulation with reality (sim-to-real) through gradient-based system identification.}

\section{Related Works}
\subsection{Soft Swimmer Muscles}
Several pioneering works have demonstrated the ability of fish-like swimming~\cite{biomimeticfish, Anderson2002ManeuveringAS}. Advances in biomimetic movement, mimicking and interacting with real fish~\cite{marchesefish, fishinteraction} has been utilized further. Fluidic actuators have been used in almost all recent fish designs. In particular, hydraulic actuators have been proposed~\cite{Katzschmann2018ExplorationFish}, which benefit from many different design and fabrication techniques like soft lithography~\cite{softlithography}, retractable pin casting~\cite{retractablepincasting} or monolithic casting with a lost-wax fabrication. Multiple swimmers using electrostatic type actuators have also been shown before~\cite{tang2021, shintake2018}. Actuators such as  dielectric elastomer actuators (DEAs) offer large actuation strains with high efficiency~\cite{DEApaper}. Because DEAs are usually in a stacked configuration with very thin dielectric and stretchable electrode layers, they are prone to electrical breakdowns and fatigue~\cite{kellaris2018}. Over the last few years HASELs have shown potential in many different applications~\cite{haselrothemund}. HASELs have high power density and are efficient, especially when recuperating high voltage power supplies are used. They are able to exert high forces~\cite{KELLARIS2019100449} and specific torque outperforming comparable electrical servo motors~\cite{kellarislsjoints}. Furthermore, HASELs offer fast actuation speeds making them well-suited for use in soft robotic systems~\cite{prosthetichasel}.

\subsection{Soft Robot Simulation}
Because soft robotic systems are continuously deforming structures, FEM is usually the only simulation method that will accurately represent the dynamics of the robot~\cite{Duriez2017FrameworkModel, Tonkens2020, Du2021DiffPD:Dynamics}. Due to the computational complexity of FEM, simplifying assumptions are often used in practice, such as Piecewise Constant Curvature models~\cite{RobertJ.Webster2010DesignReview:}, which work for only a limited range of soft robot morphologies. On the other hand, data-driven approaches have the potential to work for a wider variety of robots, however, usually require specific experimental measurements to fit the robot~\cite{Bruder2019, Haggerty2020, han2021desko}. 

In the soft robotics community, a popular framework for accurate FEM simulations is the Simulation Open Framework Architecture (SOFA)~\cite{faure:hal-00681539}. This framework is able to model highly deformable objects, and has been used for reduced-order models~\cite{Goury2018FastReduction, katzschmann2019dynamically, Tonkens2020}. In recent years, differentiable simulations have gained interest in the robotics community. Differentiable simulators allow for computation of gradients of any variable (state, model, or control) in a system with respect to a simulation objective, \textit{e.g.}, object velocity, force, distance to target. Several downstream soft robotic applications are possible due to this differentiable property, including system identification~\cite{Hu2019ChainQueen:Robotics}, trajectory optimization~\cite{Geilinger2020ADD:Contact}, motion control~\cite{qiao_differentiable_2021}, and shape optimization~\cite{ma_diffaqua_2021}. Transferring these simulations to real robots often struggle with the so-called ``sim-to-real gap'', which has been for instance overcome with system identification to correct the simulated system to match the real world parameters~\cite{Hahn2019Real2Sim:Motion, zhang_learning_2021, dubied2022sim2real}. These methods for sim-to-real were either only shown on non-actuated systems, or only on fluidic actuation, which is drastically different in terms of dynamics from electrostatic actuators.

\section{METHODOLOGY}

\subsection{Fish design and swim environment}
The swimmer’s profile is generated using a parametric polynomial adapted from~\cite{curatolo_virtual_2015}. The fish measures \SI{190}{\milli\meter} from fin to head, where the fin is \SI{30}{\milli\meter} long and the solid head is \SI{52}{\milli\meter} long. The maximum width of the fish is \SI{21}{\milli\meter}. The outer structure of the swimmer is 3D printed with a multimaterial printer described in detail in Chapter \ref{mulit}. The HASEL fish swims in canola oil because the electrostatic actuators are currently not insulated for ease of manufacturing and simulation; oil has a lower Reynolds number than water and fluids with less turbulence will be easier to simulate. The fish is connected to two custom \SI{5}{\watt} high voltage (\SI{10}{\kilo\volt}) power supplies  \cite{mitchell2019} controlled with an Arduino through a very thin conductive and electrically insulated yarn (Datastretch from Muca).

\subsection{Multi-Material Printing}\label{mulit}
\begin{figure}[!hbt]
    \centering
    \includegraphics[width=\columnwidth]{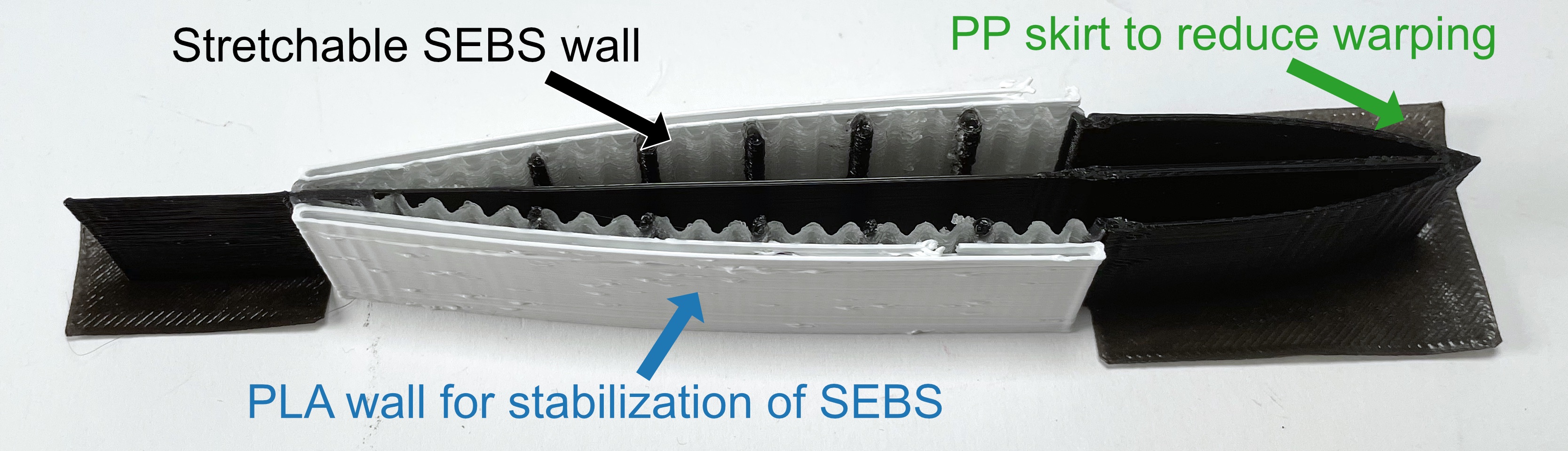}
    \caption{Fish structure directly after printing. The long continuous polypropylene (PP) wall (spine of the fish) requires brims to reduce warping. The polylactic acid (PLA) wall mechanically stabilizes the styrene-ethylene-butylene-styrene (SEBS) soft shell during printing.}
    \label{fig:fishprint}
\end{figure}

The structure of the fish is 3D printed on a tool-changing CoreXY FDM printer (adapted Rat Rig V-Core 3) using two different types of extruders. The printer can switch between two E3D Hemera extruders\footnote{https://e3d-online.com/products/e3d-hemera-direct-kit-1-75mm} for filaments and two V4 pellet extruders\footnote{https://mahor.xyz/producto/v4-pellet-extruder/}. The body of the fish is printed using a polypropylene (PP) (Shore hardness 72D) filament for the semi-rigid parts and styrene-ethylene-butylene-styrene (SEBS) pellets (Shore hardness 18A) for the soft parts. Additionally, polylactic acid (PLA) filament is used to print a stabilization wall, strengthening the SEBS parts and provides rigidity for tall prints. A 3D printing adhesive from Magigoo is used to ensure adhesion of PP and SEBS on the printbed. \Cref{fig:fishprint} shows the fish structure directly after printing with the stabilization structure and skirts still attached.

\subsection{HASEL Manufacturing}
\begin{figure}[!hbt]
    \centering
    \includegraphics[width=\columnwidth]{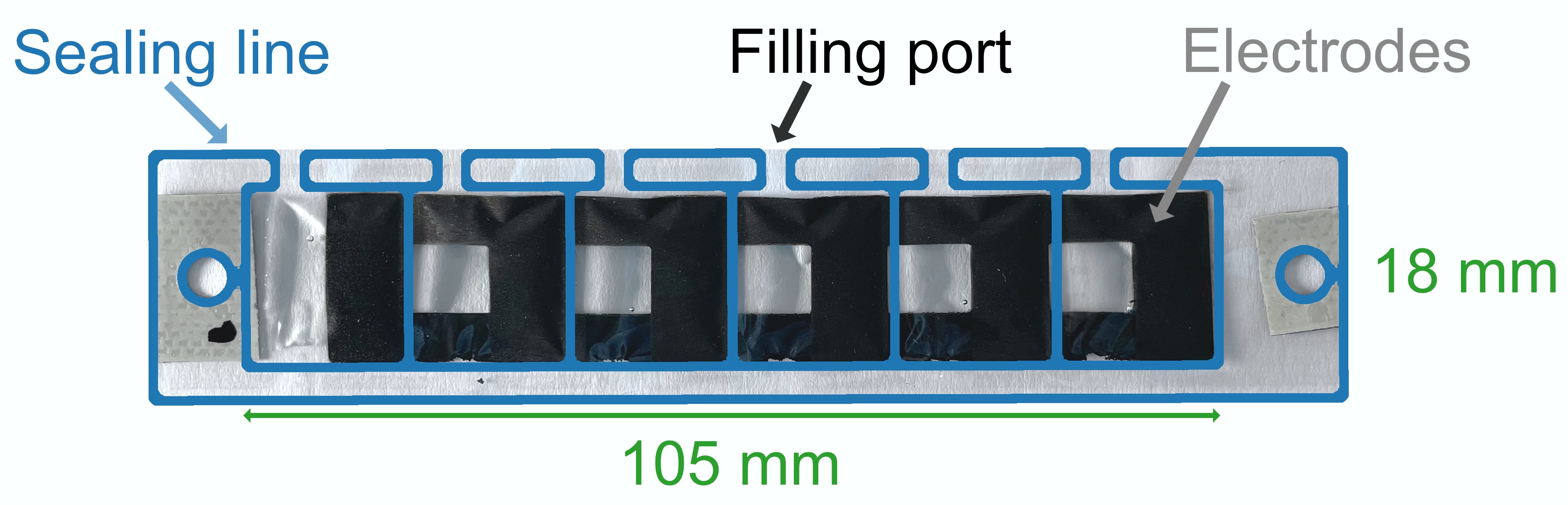}
    \caption{Finished HASEL actuator overlaid with the seal design. Each chamber has a separate filling port which is sealed with a soldering iron after filling. Two mounting holes are placed on either side to secure the actuator in the fish.}
    \label{fig:haselactuator}
\end{figure}

We adapt the method shown in \cite{mitchell2019} to manufacture the Peano HASEL actuators. We use a commercially available 3D printer (Prusa MK3S) to heat-seal the outer shell (Mylar 850, Petroplast GmbH) of the actuator. In contrast to \cite{mitchell2019} we do not use a soldering iron for the automated heat-sealing. Instead, we extrude one layer of filament onto a protective Kapton sheet (Kapton \SI{25}{\micro\meter} from RS Online) to evenly transfer heat to the Mylar films below. This avoids the need for a neoprene rubber sheet and lubrication, and makes scratching of the dielectric films less likely. Additionally, this method does not require any adaptions to the 3D printer. The actuator design shown on \Cref{fig:haselactuator} was created in Fusion 360 v2.0.11415 and exported to PrusaSlicer v2.3.1. All lines are \SI{1.5}{\milli\meter} wide and are heat-sealed twice to make the seals more durable. Electrodes are made from the carbon black based color Electrodag 502 (Gloor Instruments GmbH), which is airbrushed onto the actuator. To define the electrode area, masks are 3D printed and clamped onto the actuator using binder clips. The HASELs are filled with oil through a \SI{1.5}{\milli\meter} wide filling port, which is added to the sealing line (\Cref{fig:haselactuator}). The amount of filling liquid is chosen lower compared to other works \cite{kellaris2018} at approximately 75\% of the theoretical maximum cylindrical volume of the pouch because of the low width of the actuator pouches. This volume of dielectric oil is injected into the sample using a \SI{1}{\milli\liter} syringe, minimising the presence of air bubbles. Lastly, the filling ports are heat-sealed with a soldering iron.

\subsection{Peano HASEL characterization}
The vertical displacement of the weights is measured using a Baumer OM70-11216521 distance measurement sensor. A Trek 20/20C-HS high voltage amplifier is used to generate the high voltage signal. The low voltage input signal for the amplifier is generated by a MyDAQ from National Instruments controlled by custom LabView code. The program creates a bipolar voltage curve for the amplifier and reads the high voltage level and distance back from the high voltage amplifier and the distance measurement sensor. The force/strain performance \Cref{fig:hasel_measrument} of the actuators is generated by lifting weights three times using a bipolar voltage signal (frequency \SI{0.1}{\hertz}, duty cycle 50\%) and averaging the resulting distances.

\subsection{Simulated Swimmer Modeling}


We modeled the swimmer as a filled shape using the 5th-order parametric polynomial defined for carangiforms~\cite{curatolo2015virtual}, the same shape we used for the printed swimmer. After measuring the dimensions of the real swimmer, we chose the parametric height and length of the swimmer to be \SI{15}{\milli\meter} and \SI{160}{\milli\meter} respectively. We modified this shape by adding a more rigid spine with thickness \SI{1}{\milli\meter} that extends into a tail that is \SI{30}{\milli\meter} long. During meshing, we increased the meshing resolution near the spine within the body, not the tail, so we achieved a mean spine meshing error (distance of elements that should not belong to the spine) of \SI{0.07}{\milli\meter}. This resulted in a mesh with 1396 vertices shown in \Cref{fig:simvsreal}.

The materials we used for the 3D-printed swimmer are SEBS for the soft white material, with Young's Modulus \SI{0.65}{\mega\pascal} and density \SI{940}{\kilo\gram\per\cubic\meter}, and PP for the  rigid black material, with Young's Modulus \SI{1.33}{\giga\pascal} and density \SI{920}{\kilo\gram\per\cubic\meter}. We decided to lower the stiffness of both materials since the structure in simulation is a filled body instead of the thin-walled geometry in the real world. We modelled the stiffness of the SEBS and PP materials  with \SI{65}{\kilo\pascal} and \SI{0.13}{\mega\pascal}, respectively. \mm{We approximate the density of both materials with} \SI{900}{\kilo\gram\per\cubic\meter}. We assume that both are incompressible materials, and hence selected a Poisson's Ratio of \num{0.45}. The reasoning why we model the structure as a filled body, is due to how we define muscles. We opted for modeling the upper and lower halves of the swimmer to be contractile muscles using simplified ``Muscle Models'' (the contractile deformation can be seen in \Cref{fig:simvsreal}), as defined in~\cite{min2019softcon, Du2021DiffPD:Dynamics, dubied2022sim2real}. These simplified artificial muscles are an efficient alternative to simulating the accurate thin-wall geometry and electrostatic actuation of our swimmer, and we show that using the differentiability of the soft-body simulation we can accurately match simulation and reality. We match \mm{the artificial} muscle models\mm{, \textit{i.e.}, the contractile actuation parameters,} to data collected from the real swimmer. We will further describe this process in the next two sections and utilize the differentiability offered by our choice of differentiable simulator ``DiffPD''~\cite{Du2021DiffPD:Dynamics}.

    

\begin{figure}[!tb]
    \centering
    \begin{subfigure}{0.17\columnwidth}
        \centering
        \includegraphics[width=\columnwidth, trim={30em 45em 30em 25em},clip]{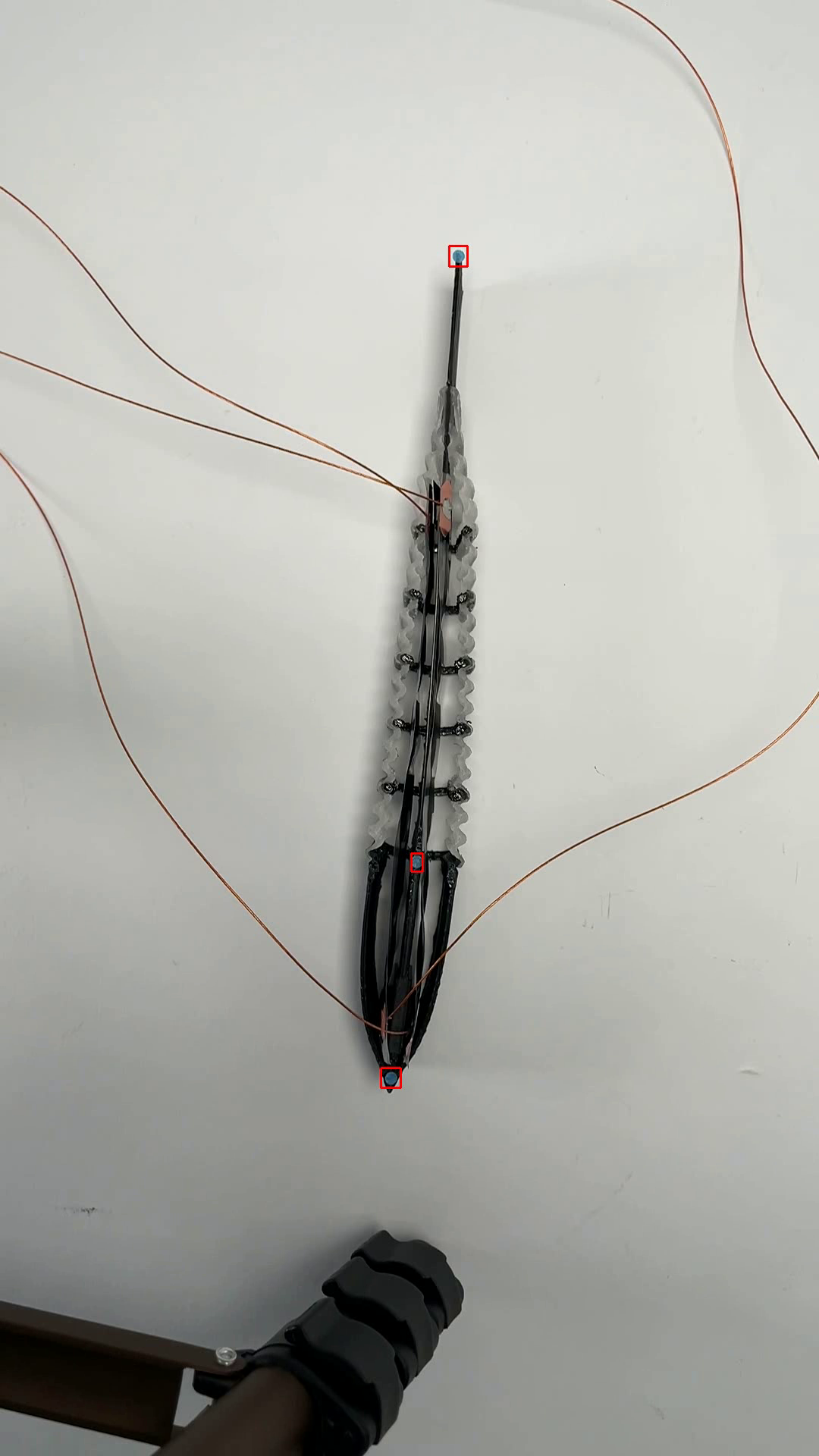}
        \caption{}
    \end{subfigure}
    \begin{subfigure}{0.17\columnwidth}
        \centering
        \includegraphics[width=\columnwidth, trim={30em 45em 30em 25em},clip]{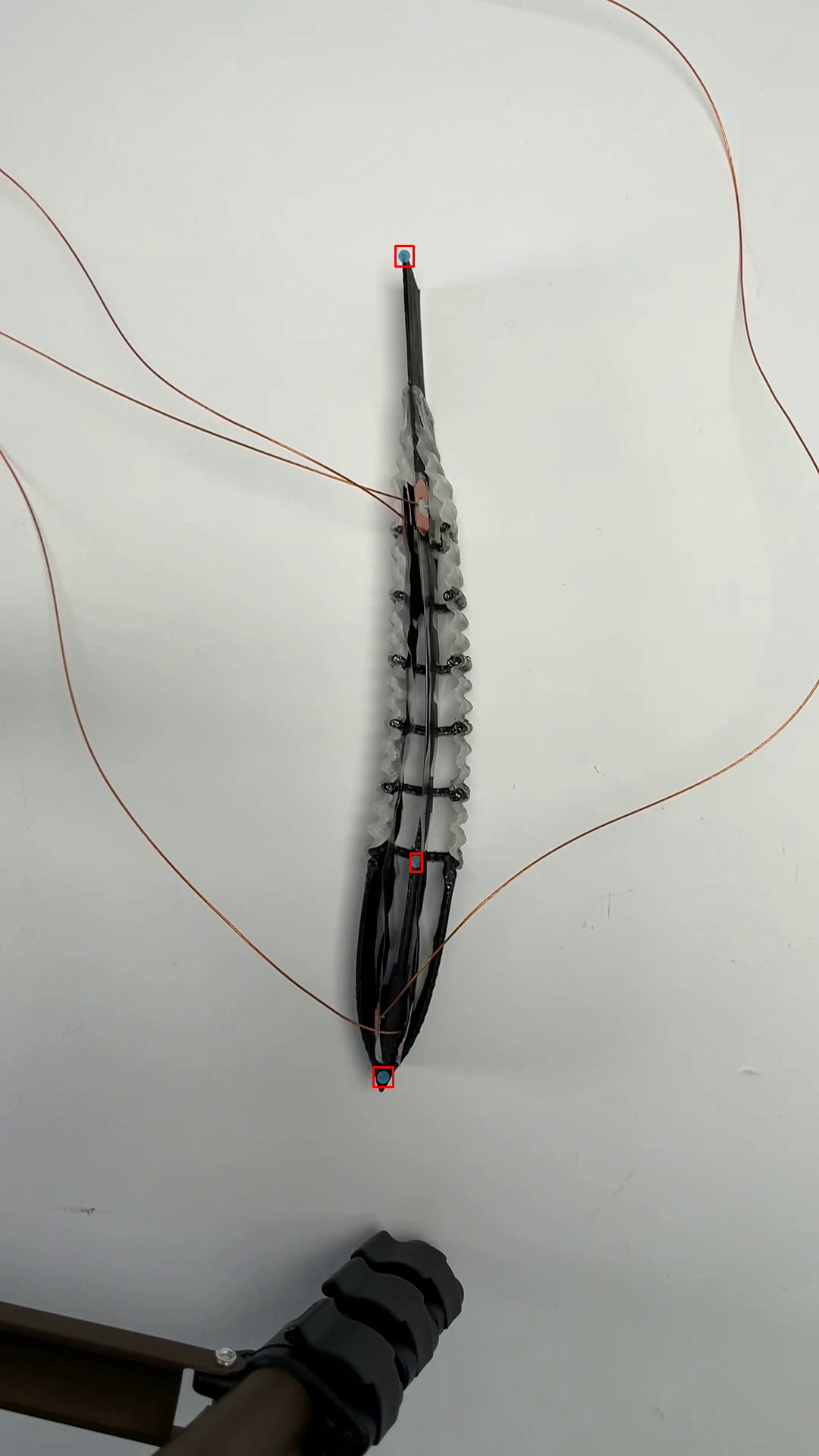}
        \caption{}
    \end{subfigure}
    \hfill
    \begin{subfigure}{0.62\columnwidth}
        \centering
        \includegraphics[width=\columnwidth]{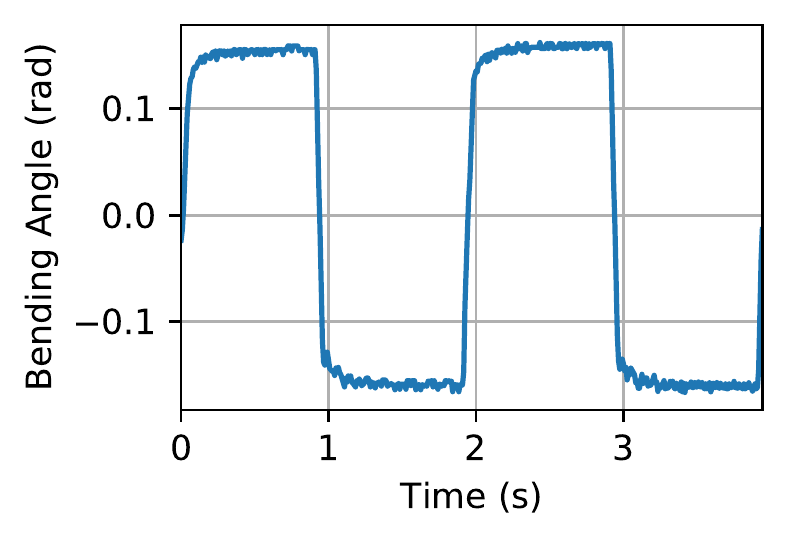}
        \caption{}
    \end{subfigure}
    \caption{Tracking of marker points on the real fish and computing the bending angle of the swimmer. (a) shows the relaxed state, and (b) the deformed state where the top muscle is actuated at \SI{5.5}{\kilo\volt}, and (c) shows two periods of HASEL actuation at \SI{0.5}{\hertz}.}
    \label{fig:anglecapture}
\end{figure}

\begin{figure}[!b]
    \centering
    \includegraphics[width=\columnwidth, trim={0em 4.5em 0em 1.8em},clip]{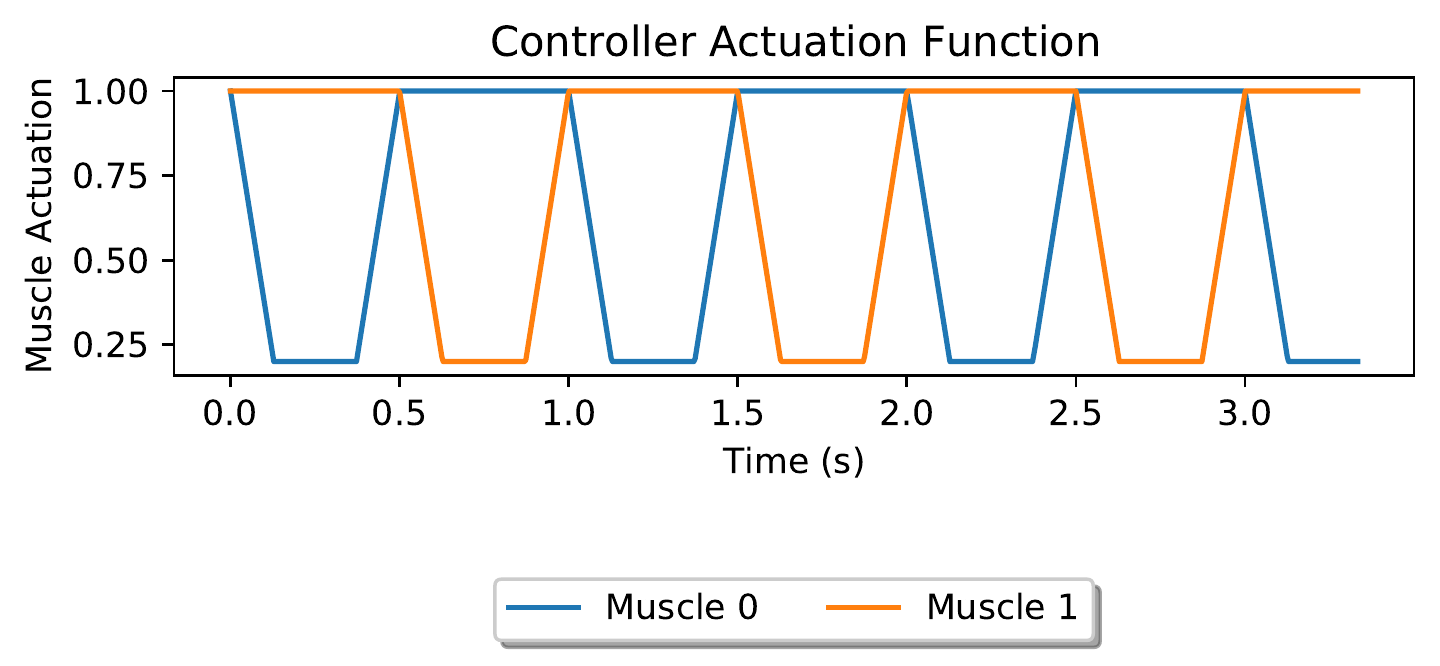}
    
    \caption{Sloped box actuation function for the simplified muscles. Top (blue) and bottom (orange) muscles are contracted sequentially. Contraction occurs with an actuation below 1.}
    \label{fig:boxactivation}
\end{figure}

\subsection{Data Collection}
\label{sec:datacollect}

To close the sim-to-real gap, we collected data from the real HASEL swimmer by actuating it using (a) varying voltages, to match the static behavior, and (b) different frequencies, to match the dynamic behavior. We collected data for voltages in $[\num{3.5}, \num{4.5}, \num{5.5}]$\SI{}{\kilo\volt} and frequencies in $[\num{0.5}, \num{2.0}, \num{3.5}]$\SI{}{\hertz}, resulting in a total of 9 datapoints for system identification. We captured the movement of the real swimmer using a camera recording from a top-down perspective at $\num{1920}\times\num{1080}$ resolution with a framerate of \SI{120}{\hertz}. We placed three blue circles on the swimmer: one on the front of its head, one in the middle of the swimmer, and lastly one on the tip of its tail.

We used the CSRT algorithm \cite{lukezic2017discriminative} to track the three marker points on the swimmer, and cut the video to the first frame where the upper muscle starts moving and the last frame after 2 periods have passed since the movement begins. We manually selected the bounding boxes for these keypoints in the starting frame, and tracked these three points over the 2 periods. We then computed the sine of the angle $\theta$ spanned by the vector pointing from head to middle $\vec{v}_{HM}$ and the vector pointing from middle to tail $\vec{v}_{MT}$ with $\sin{\theta} = (\vec{v}_{HM} \times \vec{v}_{MT}) / (\|\vec{v}_{HM}\| \, \|\vec{v}_{MT}\|)$. This bending angle will be invariant to the global motion of the swimmer, since the entire body may move due to friction with the underlying surface. Additionally, the bending angle removes the need for us to map between pixel coordinates and physical units. The sine of the angle will be positive when the top muscle is contracting, while it will be negative for the bottom muscle. We notice that in the real world the swimmer is slightly bent in its relaxed state, hence while processing the data, we subtract the average of the all bending angles to assure an as symmetric as possible behavior, such that the simulation can fit it more accurately, since it is by design symmetric. Lastly we store $\sin{\theta}$ in CSV files that we can use for system identification. \Cref{fig:anglecapture} shows how the captured data.

\subsection{Simulation System Identification}
\label{sec:sysid}

The data collected from the previous section is now compared to the simulation results, and the simulated swimmer is optimized to match reality as best as possible. We can define the exact same three keypoints on the simulated structure, and compute their angle using the same vector directions as defined previously for the real swimmer. The parameters we optimize for the simulated swimmer are (a) muscle actuation amplitude and (b) slope of the box actuation signal seen in \Cref{fig:boxactivation}. Since our simulation is differentiable, we can optimize w.r.t. these parameters given a loss objective. Our objective will be a mean squared error of simulated angle to real angle at every frame $i$ of the simulation $\frac{1}{N} \sum_{i=1}^N \| \theta_{sim}(i) - \theta_{real}(i) \|^2$. We use Adam optimizer with a learning rate of $0.05$ for the amplitude and $0.005$ for slope.

We optimized with the following assumptions: (a) given the same voltage of the real data, the muscle actuation amplitude should be the same for all actuation frequencies, and (b) the actuation slope will be the same for all samples and characterizes the HASEL swimmer (HASEL actuation + material behavior) itself. We verified the success of our system identification by testing on unseen data, namely on \SI{2.5}{\hertz} at \SI{4.25}{\kilo\volt}, \SI{4.8}{\kilo\volt}, and \SI{5.2}{\kilo\volt}. 

\begin{figure}[!b]
    \centering
    \includegraphics[width=0.8\columnwidth]{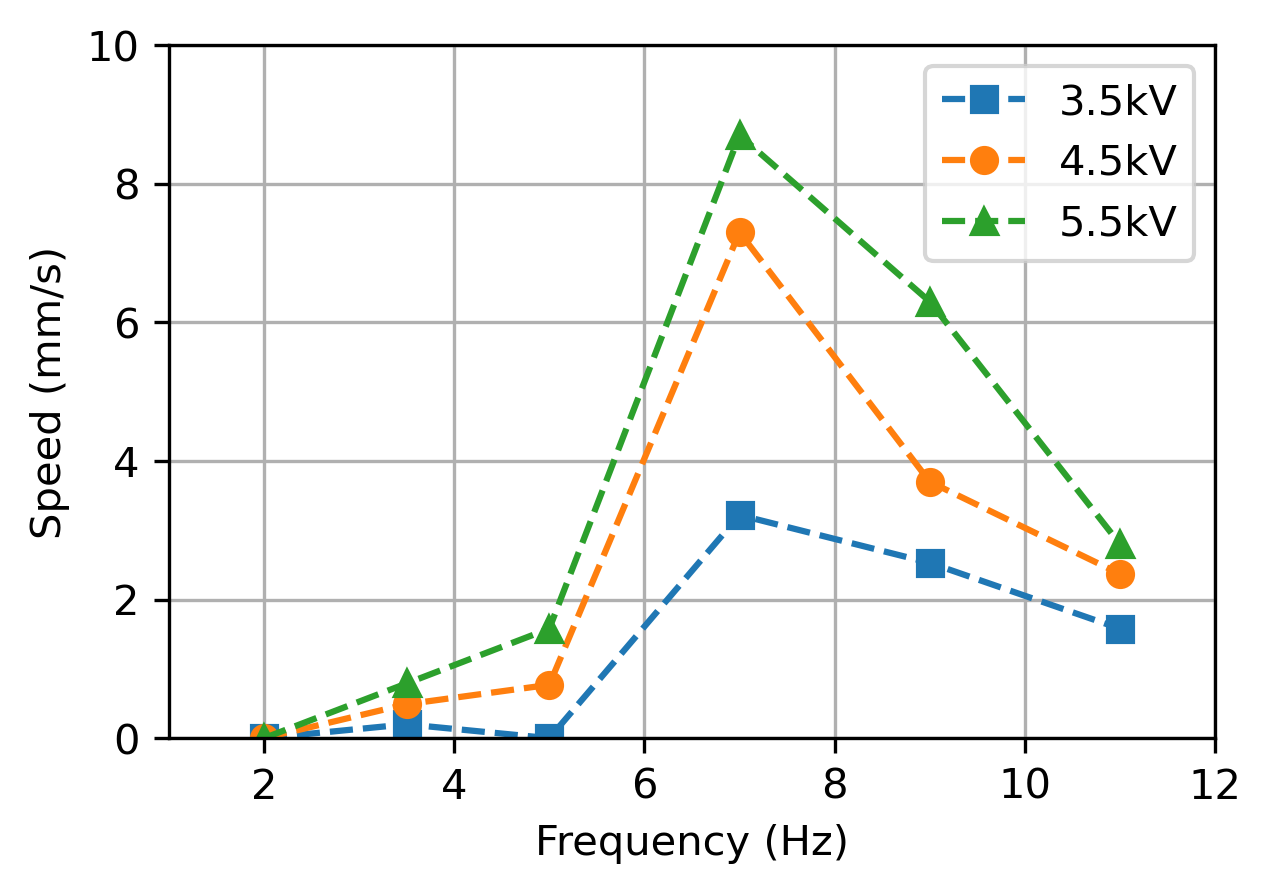}
    \caption{Speed of the swimmer in an oil tank actuated with different frequencies and voltages. At \SI{7}{\hertz} there is an optimum for locomotion speed across all voltages.}
    \label{fig:fish}
\end{figure}

\section{RESULTS}
\begin{figure}[!t]%
    \centering
    \includegraphics[width=\columnwidth]{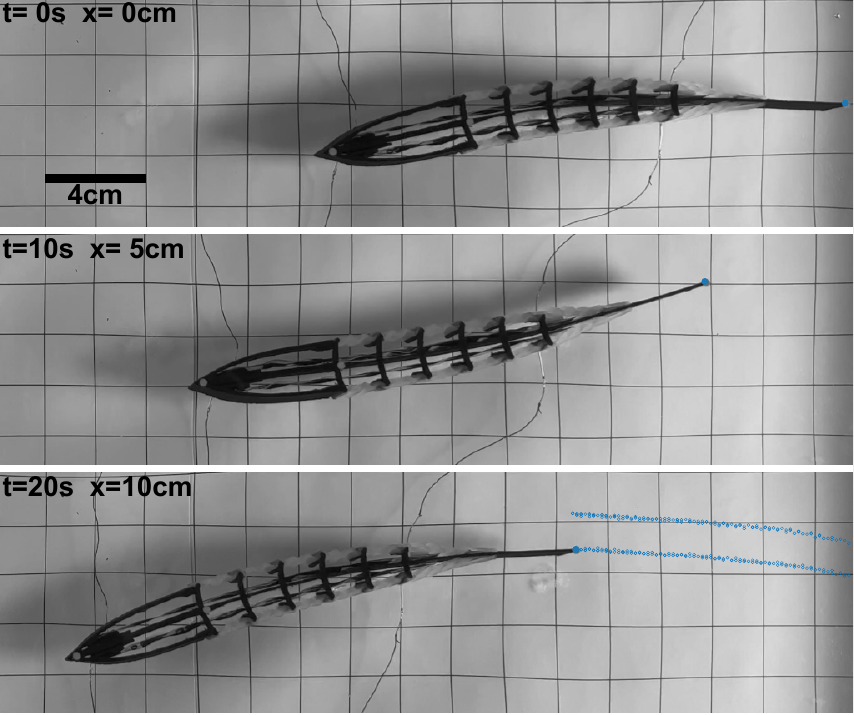}
    \caption{A time series of still images of the 3D printed fish swimming in oil at \SI{0}{\second}, \SI{10}{\second}, and \SI{20}{\second}. The swimmer traversed \SI{10}{\centi\meter} within the \SI{20}{\second} experiment. The current position of the tip of the tail fin is marked with a blue label. In the bottom image, the tail fin's maximal positions are visualized with blue lines.}
    \label{fig:swimmingFish}
\end{figure}

We submerged our swimmer into a bath of canola oil and measured the locomotion speed at different frequencies and voltages. \Cref{fig:fish} shows a local optimum for fish speed at \SI{7}{\hertz} which applies to all voltages. Above \SI{5.5}{\kilo\volt} breakdown was likely. \Cref{fig:swimmingFish} shows a typical locomotion behavior of the fish with a slight turning angle because of imperfections in actuator production and fish manufacturing. This turning angle can be compensated for by applying different voltages to each HASEL muscle. 

\begin{figure}[!b]
    \centering
    \includegraphics[width=0.8\columnwidth]{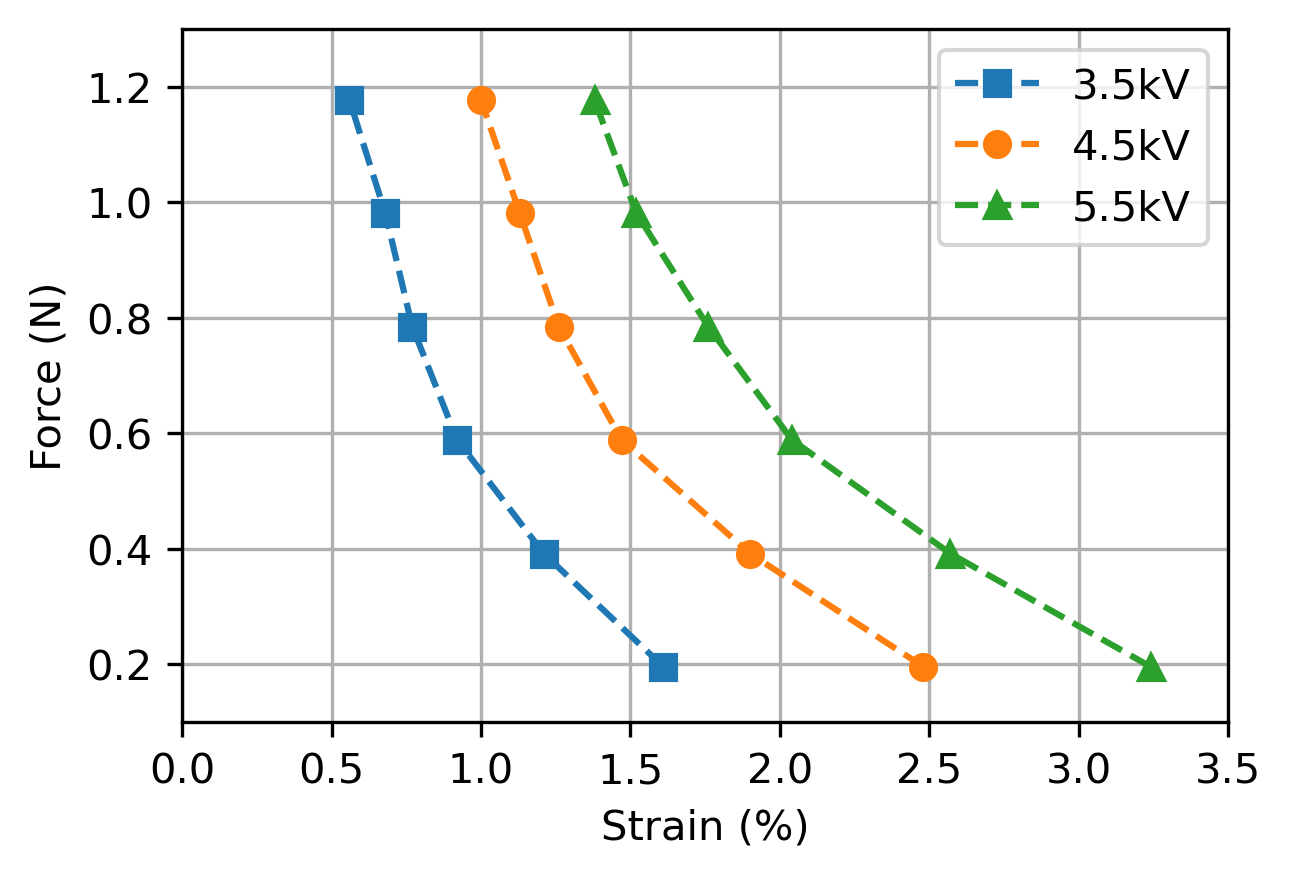}
    \caption{Force/strain performance of HASEL actuators used for actuation of the swimmer.}
    \label{fig:hasel_measrument}
\end{figure}

We investigated the performance of the Peano HASELs used for the fish. The resulting force/strain values are shown in \Cref{fig:hasel_measrument} Our actuators are strong enough to actuate the fish but significantly underperform the analytical model given by \cite{KELLARIS2019100449} for Peano HASELs with a width of \SI{18}{\milli\meter}. Our guess is that for small widths the model predictions are not accurate because edge effects dominate the zipping behavior.

\subsection{Sim-to-Real Performance}

\begin{figure}[!tb]
    \centering
    \begin{subfigure}{0.49\columnwidth}
        \includegraphics[height=8.5em]{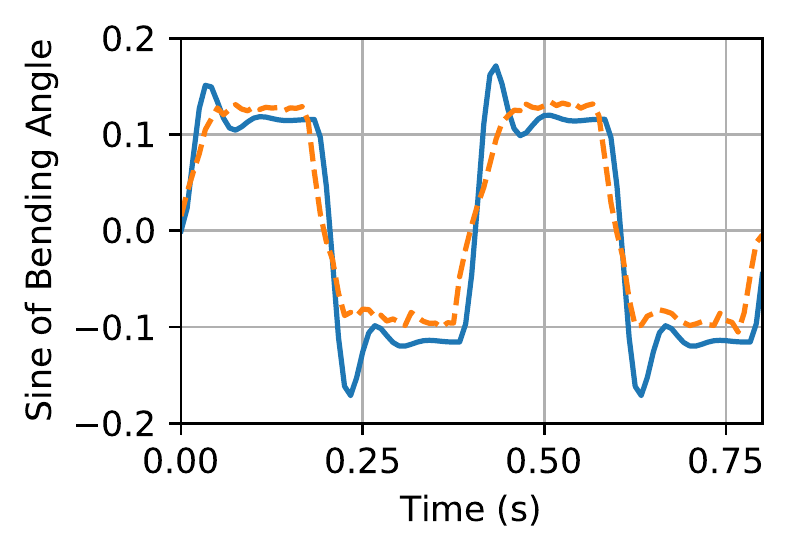}
        \caption{Voltage at \SI{4.25}{\kilo\volt}}
    \end{subfigure}
    \begin{subfigure}{0.49\columnwidth}
        \includegraphics[height=8.5em, trim={2.3em 0 0 0}, clip]{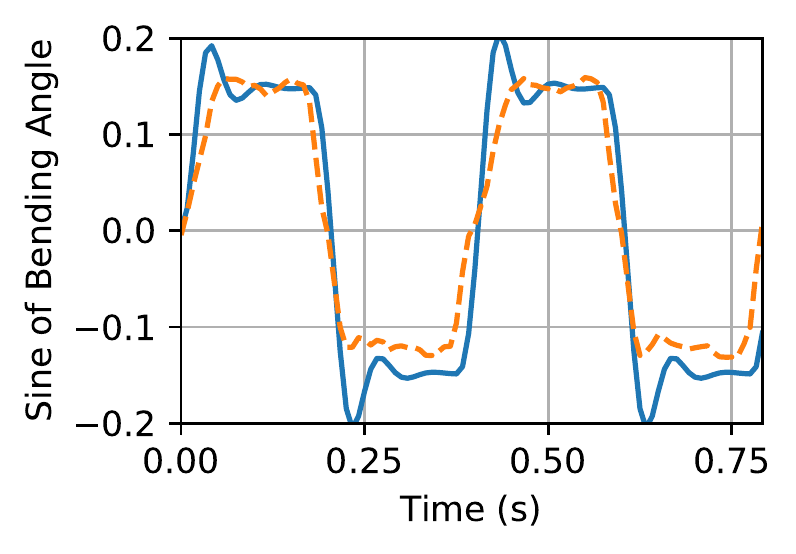}
        \caption{Voltage at \SI{5.2}{\kilo\volt}}
    \end{subfigure}
    
    \caption{The simulation results (solid blue) and real bending angle (dashed orange). We show two periods of HASEL actuation at \SI{2.5}{\hertz} for every unseen test voltage.}
    \label{fig:valangles}
\end{figure}

We optimized the muscle actuation amplitude and slope as described in \Cref{sec:sysid}, and found a resulting mean absolute angle error $\varepsilon_{\theta} =$ \num{1.553e-02} for the nine data points acquired during training. \mm{Note that the mean absolute angle error is defined by $\varepsilon_{\theta} = \frac{1}{N} \sum_{i=1}^N | \theta_{sim}(i) - \theta_{real}(i) |$.} We can show generalization of our muscle model to unseen scenarios by validating our optimized result for amplitude and slope on three voltages at a fixed frequency, all unseen to the optimization problem. We interpolate the muscle actuation from our previous data, and plug in the frequency since we assume this to be consistent between simulation and reality. The result can be seen in \Cref{fig:valangles}, and we computed a mean absolute angle error of $\varepsilon_{\theta} =$ \num{3.103e-02}. Both in amplitude and frequency this shows a reasonable generalizability that we can in future work use in downstream tasks that require the differentiability this model offers, such as control or shape optimization \cite{ma_diffaqua_2021, du2021underwater}.

\section{DISCUSSION AND FUTURE WORK}
With this work, we have shown the feasibility of using HASEL actuators as muscles for underwater robotic systems. The rapid multimaterial 3D printing approach simplifies prototyping different soft robot designs, and we have shown that simplified muscle models can be used to model HASELs for FEM simulations. We observe that the HASEL force-strain curve lends itself well to the placement and geometry in the fish. Bending the spine out of its initial straight position requires the strongest force. This equals the maximum force available at low strains. We have tested our soft robotic swimmer for various voltages and frequencies, and believe that as a next step control optimization would be greatly beneficial for showing that optimized, fast, efficient swimmer designs are possible. 

Our work exhibits several limitations, for example, the number of datapoints was too limited to assure a general model for the sim-to-real mapping. We could consider applying more sophisticated methods such as multilayer perceptrons from machine learning for learning this relationship when enough data is given. We also noted that the dynamic response, especially the damping, of DiffPD does not entirely correspond to the deformation behavior of the real swimmer. This can be compensated for explicitly \cite{dubied2022sim2real}, or we could create more complex models for our real structure.

For future work we aim to create an untethered version of the fish, incorporating new dielectric shell materials to lower the required voltages and therefore the size of internal power supplies. As a next step we also want to insulate the fish against water to make it useful for real-world applications. We also wish to optimize the fish movement and design submerged in fluid. Existing works often assume simplified hydrodynamics to allow fluid-structure interaction for underwater swimmers \cite{du2021underwater, obayashi2022soft}. As a next step, we will combine the differentiable soft body simulation used in this work with a learned hydrodynamics model \cite{wandel_learning_2021} to assure differentiability through the entire fluid-solid coupling. As far as we are aware, no one has applied soft robot optimization on real robots yet using complete, realistic hydrodynamics to describe a differentiable, multiphysical simulation.

\balance    




\section*{ACKNOWLEDGMENT}

We would like to thank Barnabas Gavin Cangan for his insightful comments during marker tracking, and Oliver Fischer for his help polishing visuals.


\bibliographystyle{IEEEtran}
\bibliography{library.bib, SRL-ETH.bib}

\begin{thebibliography}{10}
\providecommand{\url}[1]{#1}
\csname url@samestyle\endcsname
\providecommand{\newblock}{\relax}
\providecommand{\bibinfo}[2]{#2}
\providecommand{\BIBentrySTDinterwordspacing}{\spaceskip=0pt\relax}
\providecommand{\BIBentryALTinterwordstretchfactor}{4}
\providecommand{\BIBentryALTinterwordspacing}{\spaceskip=\fontdimen2\font plus
\BIBentryALTinterwordstretchfactor\fontdimen3\font minus
  \fontdimen4\font\relax}
\providecommand{\BIBforeignlanguage}[2]{{%
\expandafter\ifx\csname l@#1\endcsname\relax
\typeout{** WARNING: IEEEtran.bst: No hyphenation pattern has been}%
\typeout{** loaded for the language `#1'. Using the pattern for}%
\typeout{** the default language instead.}%
\else
\language=\csname l@#1\endcsname
\fi
#2}}
\providecommand{\BIBdecl}{\relax}
\BIBdecl

\bibitem{kuindersma_optimization-based_2016}
S.~Kuindersma, R.~Deits, M.~Fallon, A.~Valenzuela, H.~Dai, F.~Permenter,
  T.~Koolen, P.~Marion, and R.~Tedrake, ``Optimization-based locomotion
  planning, estimation, and control design for the atlas humanoid robot,''
  \emph{Auton. Robots}, vol.~40, no.~3, pp. 429--455, 2016, publisher:
  Springer.

\bibitem{miki2022learning}
T.~Miki, J.~Lee, J.~Hwangbo, L.~Wellhausen, V.~Koltun, and M.~Hutter,
  ``Learning robust perceptive locomotion for quadrupedal robots in the wild,''
  \emph{Science Robotics}, vol.~7, no.~62, p. eabk2822, 2022.

\bibitem{rus_design_2015}
\BIBentryALTinterwordspacing
D.~Rus and M.~T. Tolley, ``\BIBforeignlanguage{en}{Design, fabrication and
  control of soft robots},'' \emph{\BIBforeignlanguage{en}{Nature}}, vol. 521,
  no. 7553, pp. 467--475, May 2015, number: 7553 Publisher: Nature Publishing
  Group. [Online]. Available: \url{https://www.nature.com/articles/nature14543}
\BIBentrySTDinterwordspacing

\bibitem{hawkes_hard_2021}
\BIBentryALTinterwordspacing
E.~W. Hawkes, C.~Majidi, and M.~T. Tolley, ``\BIBforeignlanguage{en}{Hard
  questions for soft robotics},'' \emph{\BIBforeignlanguage{en}{Science
  Robotics}}, vol.~6, no.~53, Apr. 2021, publisher: Science Robotics Section:
  Viewpoint. [Online]. Available:
  \url{https://robotics.sciencemag.org/content/6/53/eabg6049}
\BIBentrySTDinterwordspacing

\bibitem{morin_camouflage_2012}
S.~A. Morin, R.~F. Shepherd, S.~W. Kwok, A.~A. Stokes, A.~Nemiroski, and G.~M.
  Whitesides, ``Camouflage and display for soft machines,'' \emph{Science},
  vol. 337, no. 6096, pp. 828--832, 2012, publisher: American Association for
  the Advancement of Science.

\bibitem{miriyev_soft_2017}
\BIBentryALTinterwordspacing
A.~Miriyev, K.~Stack, and H.~Lipson, ``\BIBforeignlanguage{en}{Soft material
  for soft actuators},'' \emph{\BIBforeignlanguage{en}{Nature Communications}},
  vol.~8, no.~1, p. 596, Sep. 2017, number: 1 Publisher: Nature Publishing
  Group. [Online]. Available:
  \url{https://www.nature.com/articles/s41467-017-00685-3}
\BIBentrySTDinterwordspacing

\bibitem{sachyani_keneth_multi-material_2020}
\BIBentryALTinterwordspacing
E.~Sachyani~Keneth, R.~Lieberman, M.~Rednor, G.~Scalet, F.~Auricchio, and
  S.~Magdassi, ``\BIBforeignlanguage{en}{Multi-{Material} {3D} {Printed}
  {Shape} {Memory} {Polymer} with {Tunable} {Melting} and {Glass} {Transition}
  {Temperature} {Activated} by {Heat} or {Light}},''
  \emph{\BIBforeignlanguage{en}{Polymers}}, vol.~12, no.~3, p. 710, Mar. 2020,
  number: 3 Publisher: Multidisciplinary Digital Publishing Institute.
  [Online]. Available: \url{https://www.mdpi.com/2073-4360/12/3/710}
\BIBentrySTDinterwordspacing

\bibitem{ilami_materials_2020}
M.~Ilami, H.~Bagheri, R.~Ahmed, E.~O. Skowronek, and H.~Marvi, ``Materials,
  {Actuators}, and {Sensors} for {Soft} {Bioinspired} {Robots},''
  \emph{Advanced Materials}, p. 2003139, 2020, publisher: Wiley Online Library.

\bibitem{berlinger_fish-like_2021}
\BIBentryALTinterwordspacing
F.~Berlinger, M.~Saadat, H.~Haj-Hariri, G.~V. Lauder, and R.~Nagpal,
  ``\BIBforeignlanguage{en}{Fish-like three-dimensional swimming with an
  autonomous, multi-fin, and biomimetic robot},''
  \emph{\BIBforeignlanguage{en}{Bioinspiration \& Biomimetics}}, vol.~16,
  no.~2, p. 026018, Feb. 2021, publisher: IOP Publishing. [Online]. Available:
  \url{https://doi.org/10.1088/1748-3190/abd013}
\BIBentrySTDinterwordspacing

\bibitem{costa_design_2020}
\BIBentryALTinterwordspacing
D.~Costa, G.~Palmieri, M.-C. Palpacelli, D.~Scaradozzi, and M.~Callegari,
  ``\BIBforeignlanguage{en}{Design of a {Carangiform} {Swimming} {Robot}
  through a {Multiphysics} {Simulation} {Environment}},''
  \emph{\BIBforeignlanguage{en}{Biomimetics}}, vol.~5, no.~4, p.~46, Dec. 2020,
  number: 4 Publisher: Multidisciplinary Digital Publishing Institute.
  [Online]. Available: \url{https://www.mdpi.com/2313-7673/5/4/46}
\BIBentrySTDinterwordspacing

\bibitem{li_self-powered_2021}
\BIBentryALTinterwordspacing
G.~Li, X.~Chen, F.~Zhou, Y.~Liang, Y.~Xiao, X.~Cao, Z.~Zhang, M.~Zhang, B.~Wu,
  S.~Yin, Y.~Xu, H.~Fan, Z.~Chen, W.~Song, W.~Yang, B.~Pan, J.~Hou, W.~Zou,
  S.~He, X.~Yang, G.~Mao, Z.~Jia, H.~Zhou, T.~Li, S.~Qu, Z.~Xu, Z.~Huang,
  Y.~Luo, T.~Xie, J.~Gu, S.~Zhu, and W.~Yang,
  ``\BIBforeignlanguage{en}{Self-powered soft robot in the {Mariana}
  {Trench}},'' \emph{\BIBforeignlanguage{en}{Nature}}, vol. 591, no. 7848, pp.
  66--71, Mar. 2021, number: 7848 Publisher: Nature Publishing Group. [Online].
  Available: \url{https://www.nature.com/articles/s41586-020-03153-z}
\BIBentrySTDinterwordspacing

\bibitem{scaradozzi_bcf_2017}
\BIBentryALTinterwordspacing
D.~Scaradozzi, G.~Palmieri, D.~Costa, and A.~Pinelli,
  ``\BIBforeignlanguage{en}{{BCF} swimming locomotion for autonomous underwater
  robots: a review and a novel solution to improve control and efficiency},''
  \emph{\BIBforeignlanguage{en}{Ocean Engineering}}, vol. 130, pp. 437--453,
  Jan. 2017. [Online]. Available:
  \url{https://www.sciencedirect.com/science/article/pii/S0029801816305613}
\BIBentrySTDinterwordspacing

\bibitem{trethewey_earths_2020}
\BIBentryALTinterwordspacing
L.~Trethewey, ``\BIBforeignlanguage{en}{Earth's final frontier: the global race
  to map the entire ocean floor},'' Jun. 2020, section: Environment. [Online].
  Available:
  \url{http://www.theguardian.com/environment/2020/jun/30/earths-final-frontier-the-global-race-to-map-the-entire-ocean-floor}
\BIBentrySTDinterwordspacing

\bibitem{krause_interactive_2011}
J.~Krause, A.~F.~T. Winfield, and J.-L. Deneubourg,
  ``\BIBforeignlanguage{en}{Interactive robots in experimental biology},''
  \emph{\BIBforeignlanguage{en}{Trends Ecol. Evol.}}, vol.~26, no.~7, pp.
  369--375, Jul. 2011.

\bibitem{marx_sargassum_2021}
\BIBentryALTinterwordspacing
U.~C. Marx, J.~Roles, and B.~Hankamer, ``\BIBforeignlanguage{en}{Sargassum
  blooms in the {Atlantic} {Ocean} – {From} a burden to an asset},''
  \emph{\BIBforeignlanguage{en}{Algal Research}}, vol.~54, p. 102188, Apr.
  2021. [Online]. Available:
  \url{https://www.sciencedirect.com/science/article/pii/S2211926421000072}
\BIBentrySTDinterwordspacing

\bibitem{gomez_jr_size_2017}
\BIBentryALTinterwordspacing
J.~E.~A. Gomez~Jr., ``\BIBforeignlanguage{en}{The size of cities: {A} synthesis
  of multi-disciplinary perspectives on the global megalopolis},''
  \emph{\BIBforeignlanguage{en}{Progress in Planning}}, vol. 116, pp. 1--29,
  Aug. 2017. [Online]. Available:
  \url{https://www.sciencedirect.com/science/article/pii/S0305900616300113}
\BIBentrySTDinterwordspacing

\bibitem{marras_fish_2012}
S.~Marras and M.~Porfiri, ``Fish and robots swimming together: attraction
  towards the robot demands biomimetic locomotion,'' \emph{Journal of The Royal
  Society …}, vol.~9, no.~73, pp. 1856--1868, Aug. 2012.

\bibitem{butail_fishrobot_2015}
S.~Butail, N.~Abaid, S.~Macrì, and M.~Porfiri,
  ``\BIBforeignlanguage{en}{Fish–{Robot} {Interactions}: {Robot} {Fish} in
  {Animal} {Behavioral} {Studies}},'' in \emph{\BIBforeignlanguage{en}{Robot
  {Fish}}}, ser. Springer {Tracts} in {Mechanical} {Engineering}.\hskip 1em
  plus 0.5em minus 0.4em\relax Springer, Berlin, Heidelberg, 2015, pp.
  359--377.

\bibitem{aracri_soft_2021}
\BIBentryALTinterwordspacing
S.~Aracri, F.~Giorgio-Serchi, G.~Suaria, M.~E. Sayed, M.~P. Nemitz, S.~Mahon,
  and A.~A. Stokes, ``Soft {Robots} for {Ocean} {Exploration} and {Offshore}
  {Operations}: {A} {Perspective},'' \emph{Soft Robotics}, Jan. 2021,
  publisher: Mary Ann Liebert, Inc., publishers. [Online]. Available:
  \url{https://www.liebertpub.com/doi/full/10.1089/soro.2020.0011}
\BIBentrySTDinterwordspacing

\bibitem{biomimeticfish}
J.~Yu, M.~Tan, S.~Wang, and E.~Chen, ``Development of a biomimetic robotic fish
  and its control algorithm,'' \emph{IEEE Transactions on Systems, Man, and
  Cybernetics, Part B (Cybernetics)}, vol.~34, no.~4, pp. 1798--1810, 2004.

\bibitem{Anderson2002ManeuveringAS}
J.~M. Anderson and N.~K. Chhabra, ``Maneuvering and stability performance of a
  robotic tuna1,'' in \emph{Integrative and comparative biology}, 2002.

\bibitem{marchesefish}
\BIBentryALTinterwordspacing
A.~D. Marchese, C.~D. Onal, and D.~Rus, ``Autonomous soft robotic fish capable
  of escape maneuvers using fluidic elastomer actuators,'' \emph{Soft
  Robotics}, vol.~1, no.~1, pp. 75--87, 2014, pMID: 27625912. [Online].
  Available: \url{https://doi.org/10.1089/soro.2013.0009}
\BIBentrySTDinterwordspacing

\bibitem{fishinteraction}
S.~Marras and M.~Porfiri, ``Fish and robots swimming together: Attraction
  towards the robot demands biomimetic locomotion,'' \emph{Journal of the Royal
  Society, Interface / the Royal Society}, vol.~9, pp. 1856--68, 02 2012.

\bibitem{Katzschmann2018ExplorationFish}
\BIBentryALTinterwordspacing
R.~K. Katzschmann, J.~DelPreto, R.~MacCurdy, and D.~Rus, ``{Exploration of
  underwater life with an acoustically controlled soft robotic fish},''
  \emph{Science Robotics}, vol.~3, no.~16, p. 3449, 3 2018. [Online].
  Available: \url{http://robotics.sciencemag.org/}
\BIBentrySTDinterwordspacing

\bibitem{softlithography}
\BIBentryALTinterwordspacing
Y.~Xia and G.~M. Whitesides, ``Soft lithography,'' \emph{Annual Review of
  Materials Science}, vol.~28, no.~1, pp. 153--184, 1998. [Online]. Available:
  \url{https://doi.org/10.1146/annurev.matsci.28.1.153}
\BIBentrySTDinterwordspacing

\bibitem{retractablepincasting}
\BIBentryALTinterwordspacing
A.~D. Marchese, R.~K. Katzschmann, and D.~Rus, ``A recipe for soft fluidic
  elastomer robots,'' \emph{Soft Robotics}, vol.~2, no.~1, pp. 7--25, 2015,
  pMID: 27625913. [Online]. Available:
  \url{https://doi.org/10.1089/soro.2014.0022}
\BIBentrySTDinterwordspacing

\bibitem{tang2021}
\BIBentryALTinterwordspacing
W.~Tang, Y.~Lin, C.~Zhang, Y.~Liang, J.~Wang, W.~Wang, C.~Ji, M.~Zhou, H.~Yang,
  and J.~Zou, ``Self-contained soft electrofluidic actuators,'' \emph{Science
  Advances}, vol.~7, no.~34, p. eabf8080, 2021. [Online]. Available:
  \url{https://www.science.org/doi/abs/10.1126/sciadv.abf8080}
\BIBentrySTDinterwordspacing

\bibitem{shintake2018}
\BIBentryALTinterwordspacing
J.~Shintake, V.~Cacucciolo, H.~Shea, and D.~Floreano, ``Soft biomimetic fish
  robot made of dielectric elastomer actuators,'' \emph{Soft Robotics}, vol.~5,
  no.~4, pp. 466--474, 2018, pMID: 29957131. [Online]. Available:
  \url{https://doi.org/10.1089/soro.2017.0062}
\BIBentrySTDinterwordspacing

\bibitem{DEApaper}
R.~Pelrine, R.~Kornbluh, Q.~Pei, and J.~Joseph, ``High-speed electrically
  actuated elastomers with strain greater than 100\%,'' \emph{Science (New
  York, N.Y.)}, vol. 287, no. 5454, p. 836–839, 2000.

\bibitem{kellaris2018}
\BIBentryALTinterwordspacing
N.~Kellaris, V.~G. Venkata, G.~M. Smith, S.~K. Mitchell, and C.~Keplinger,
  ``Peano-hasel actuators: Muscle-mimetic, electrohydraulic transducers that
  linearly contract on activation,'' \emph{Science Robotics}, vol.~3, no.~14,
  p. eaar3276, 2018. [Online]. Available:
  \url{https://www.science.org/doi/abs/10.1126/scirobotics.aar3276}
\BIBentrySTDinterwordspacing

\bibitem{haselrothemund}
\BIBentryALTinterwordspacing
P.~Rothemund, N.~Kellaris, S.~K. Mitchell, E.~Acome, and C.~Keplinger, ``Hasel
  artificial muscles for a new generation of lifelike robots—recent progress
  and future opportunities,'' \emph{Advanced Materials}, vol.~33, no.~19, p.
  2003375, 2021. [Online]. Available:
  \url{https://onlinelibrary.wiley.com/doi/abs/10.1002/adma.202003375}
\BIBentrySTDinterwordspacing

\bibitem{KELLARIS2019100449}
\BIBentryALTinterwordspacing
N.~Kellaris, V.~G. Venkata, P.~Rothemund, and C.~Keplinger, ``An analytical
  model for the design of peano-hasel actuators with drastically improved
  performance,'' \emph{Extreme Mechanics Letters}, vol.~29, p. 100449, 2019.
  [Online]. Available:
  \url{https://www.sciencedirect.com/science/article/pii/S2352431619300173}
\BIBentrySTDinterwordspacing

\bibitem{kellarislsjoints}
\BIBentryALTinterwordspacing
N.~Kellaris, P.~Rothemund, Y.~Zeng, S.~K. Mitchell, G.~M. Smith, K.~Jayaram,
  and C.~Keplinger, ``Spider-inspired electrohydraulic actuators for fast,
  soft-actuated joints,'' \emph{Advanced Science}, vol.~8, no.~14, p. 2100916,
  2021. [Online]. Available:
  \url{https://onlinelibrary.wiley.com/doi/abs/10.1002/advs.202100916}
\BIBentrySTDinterwordspacing

\bibitem{prosthetichasel}
\BIBentryALTinterwordspacing
Z.~Yoder, N.~Kellaris, C.~Chase-Markopoulou, D.~Ricken, S.~K. Mitchell, M.~B.
  Emmett, R.~F.~f. Weir, J.~Segil, and C.~Keplinger, ``Design of a high-speed
  prosthetic finger driven by peano-hasel actuators,'' \emph{Frontiers in
  Robotics and AI}, vol.~7, 2020. [Online]. Available:
  \url{https://www.frontiersin.org/article/10.3389/frobt.2020.586216}
\BIBentrySTDinterwordspacing

\bibitem{Duriez2017FrameworkModel}
C.~Duriez, E.~Coevoet, F.~Largilliere, T.~Morales-Bieze, Z.~Zhang,
  M.~Sanz-Lopez, B.~Carrez, D.~Marchal, O.~Goury, and J.~Dequidt, ``{Framework
  for online simulation of soft robots with optimization-based inverse
  model},'' in \emph{2016 IEEE International Conference on Simulation,
  Modeling, and Programming for Autonomous Robots, SIMPAR 2016}.\hskip 1em plus
  0.5em minus 0.4em\relax Institute of Electrical and Electronics Engineers
  Inc., 2 2017, pp. 111--118.

\bibitem{Tonkens2020}
S.~Tonkens, J.~Lorenzetti, and M.~Pavone, ``Soft robot optimal control via
  reduced order finite element models,'' in \emph{2021 IEEE International
  Conference on Robotics and Automation (ICRA)}, 2021, pp. 12\,010--12\,016.

\bibitem{Du2021DiffPD:Dynamics}
T.~Du, K.~Wu, P.~Ma, S.~Wah, A.~Spielberg, D.~Rus, and W.~Matusik, ``Diffpd:
  Differentiable projective dynamics,'' \emph{ACM Transactions on Graphics},
  vol.~41, no.~2, nov 2021.

\bibitem{RobertJ.Webster2010DesignReview:}
I.~Robert J.~Webster and B.~A. Jones, ``Design and kinematic modeling of
  constant curvature continuum robots: A review,'' \emph{The International
  Journal of Robotics Research}, vol.~29, no.~13, pp. 1661--1683, 2010.

\bibitem{Bruder2019}
D.~Bruder, B.~Gillespie, C.~David~Remy, and R.~Vasudevan, ``Modeling and
  control of soft robots using the koopman operator and model predictive
  control,'' \emph{arXiv: 1902.02827}, 2019.

\bibitem{Haggerty2020}
D.~A. Haggerty, M.~J. Banks, P.~C. Curtis, I.~Mezić, and E.~W. Hawkes,
  ``Modeling, reduction, and control of a helically actuated inertial soft
  robotic arm via the koopman operator,'' \emph{arXiv: 2011.07939}, 2020.

\bibitem{han2021desko}
M.~Han, J.~Euler-Rolle, and R.~K. Katzschmann, ``Desko: Stability-assured
  robust control with a deep stochastic koopman operator,'' in
  \emph{International Conference on Learning Representations}, 2021.

\bibitem{faure:hal-00681539}
\BIBentryALTinterwordspacing
F.~Faure, C.~Duriez, H.~Delingette, J.~Allard, B.~Gilles, S.~Marchesseau,
  H.~Talbot, H.~Courtecuisse, G.~Bousquet, I.~Peterlik, and S.~Cotin, ``{SOFA:
  A Multi-Model Framework for Interactive Physical Simulation},'' in
  \emph{{Soft Tissue Biomechanical Modeling for Computer Assisted Surgery}},
  ser. Studies in Mechanobiology, Tissue Engineering and Biomaterials,
  Y.~Payan, Ed.\hskip 1em plus 0.5em minus 0.4em\relax {Springer}, Jun. 2012,
  vol.~11, pp. 283--321. [Online]. Available:
  \url{https://hal.inria.fr/hal-00681539}
\BIBentrySTDinterwordspacing

\bibitem{Goury2018FastReduction}
O.~Goury and C.~Duriez, ``Fast, generic, and reliable control and simulation of
  soft robots using model order reduction,'' \emph{IEEE Transactions on
  Robotics}, vol.~34, no.~6, pp. 1565--1576, 12 2018.

\bibitem{katzschmann2019dynamically}
R.~K. Katzschmann, M.~Thieffry, O.~Goury, A.~Kruszewski, T.-M. Guerra,
  C.~Duriez, and D.~Rus, ``Dynamically closed-loop controlled soft robotic arm
  using a reduced order finite element model with state observer,'' in
  \emph{2019 2nd IEEE International Conference on Soft Robotics
  (RoboSoft)}.\hskip 1em plus 0.5em minus 0.4em\relax IEEE, 2019, pp. 717--724.

\bibitem{Hu2019ChainQueen:Robotics}
Y.~Hu, J.~Liu, A.~Spielberg, J.~B. Tenenbaum, W.~T. Freeman, J.~Wu, D.~Rus, and
  W.~Matusik, ``Chainqueen: A real-time differentiable physical simulator for
  soft robotics,'' \emph{Proceedings - IEEE International Conference on
  Robotics and Automation}, vol. 2019-May, pp. 6265--6271, 5 2019.

\bibitem{Geilinger2020ADD:Contact}
M.~Geilinger, D.~Hahn, J.~Zehnder, M.~B{\"{a}}cher, B.~Thomaszewski, and
  S.~Coros, ``Add: Analytically differentiable dynamics for multi-body
  systemswith frictional contact,'' \emph{ACM Transactions on Graphics (TOG)},
  vol.~39, no.~6, p.~15, 11 2020.

\bibitem{qiao_differentiable_2021}
Y.-L. Qiao, J.~Liang, V.~Koltun, and M.~C. Lin, ``Differentiable {Simulation}
  of {Soft} {Multi}-body {Systems},'' in \emph{Conference on {Neural}
  {Information} {Processing} {Systems} ({NeurIPS})}, 2021.

\bibitem{ma_diffaqua_2021}
\BIBentryALTinterwordspacing
P.~Ma, T.~Du, J.~Z. Zhang, K.~Wu, A.~Spielberg, R.~K. Katzschmann, and
  W.~Matusik, ``{DiffAqua}: {A} {Differentiable} {Computational} {Design}
  {Pipeline} for {Soft} {Underwater} {Swimmers} with {Shape} {Interpolation},''
  in \emph{{arXiv}:2104.00837 [cs]}, Apr. 2021, arXiv: 2104.00837. [Online].
  Available: \url{http://arxiv.org/abs/2104.00837}
\BIBentrySTDinterwordspacing

\bibitem{Hahn2019Real2Sim:Motion}
D.~Hahn, P.~Banzet, J.~M. Bern, and S.~Coros, ``Real2sim: Visco-elastic
  parameter estimation from dynamic motion,'' \emph{ACM Transactions on
  Graphics (TOG)}, vol.~38, no.~6, 11 2019.

\bibitem{zhang_learning_2021}
\BIBentryALTinterwordspacing
J.~Z. Zhang, Y.~Zhang, P.~Ma, E.~Nava, T.~Du, P.~Arm, W.~Matusik, and R.~K.
  Katzschmann, ``Learning {Material} {Parameters} and {Hydrodynamics} of {Soft}
  {Robotic} {Fish} via {Differentiable} {Simulation},'' \emph{arXiv:2109.14855
  [cs]}, Sep. 2021, arXiv: 2109.14855. [Online]. Available:
  \url{http://arxiv.org/abs/2109.14855}
\BIBentrySTDinterwordspacing

\bibitem{dubied2022sim2real}
M.~Dubied, M.~Y. Michelis, A.~Spielberg, and R.~K. Katzschmann, ``Sim-to-real
  for soft robots using differentiable fem: Recipes for meshing, damping, and
  actuation,'' \emph{IEEE Robotics and Automation Letters}, pp. 1--1, 2022.

\bibitem{curatolo_virtual_2015}
M.~Curatolo and L.~Teresi, ``The virtual aquarium: simulations of fish
  swimming,'' in \emph{Proc. {European} {COMSOL} {Conference}}, 2015.

\bibitem{mitchell2019}
\BIBentryALTinterwordspacing
S.~K. Mitchell, X.~Wang, E.~Acome, T.~Martin, K.~Ly, N.~Kellaris, V.~G.
  Venkata, and C.~Keplinger, ``An easy-to-implement toolkit to create versatile
  and high-performance hasel actuators for untethered soft robots,''
  \emph{Advanced Science}, vol.~6, no.~14, p. 1900178, 2019. [Online].
  Available:
  \url{https://onlinelibrary.wiley.com/doi/abs/10.1002/advs.201900178}
\BIBentrySTDinterwordspacing

\bibitem{curatolo2015virtual}
M.~Curatolo and L.~Teresi, ``The virtual aquarium: simulations of fish
  swimming,'' in \emph{Proc. European COMSOL Conference}, 2015.

\bibitem{min2019softcon}
S.~Min, J.~Won, S.~Lee, J.~Park, and J.~Lee, ``{Softcon: Simulation and control
  of soft-bodied animals with biomimetic actuators},'' \emph{ACM Transactions
  on Graphics (TOG)}, vol.~38, no.~6, pp. 1--12, 2019.

\bibitem{lukezic2017discriminative}
A.~Lukezic, T.~Vojir, L.~ˇCehovin~Zajc, J.~Matas, and M.~Kristan,
  ``Discriminative correlation filter with channel and spatial reliability,''
  in \emph{Proceedings of the IEEE conference on computer vision and pattern
  recognition}, 2017, pp. 6309--6318.

\bibitem{du2021underwater}
T.~Du, J.~Hughes, S.~Wah, W.~Matusik, and D.~Rus, ``Underwater soft robot
  modeling and control with differentiable simulation,'' \emph{IEEE Robotics
  and Automation Letters}, vol.~6, no.~3, pp. 4994--5001, 2021.

\bibitem{obayashi2022soft}
N.~Obayashi, C.~Bosio, and J.~Hughes, ``Soft passive swimmer optimization: From
  simulation to reality using data-driven transformation,'' in \emph{5th
  IEEE-RAS International Conference on Soft Robotics (RoboSoft 2022)}, no.
  CONF, 2022.

\bibitem{wandel_learning_2021}
\BIBentryALTinterwordspacing
N.~Wandel, M.~Weinmann, and R.~Klein, ``\BIBforeignlanguage{en}{Learning
  {Incompressible} {Fluid} {Dynamics} from {Scratch} -- {Towards} {Fast},
  {Differentiable} {Fluid} {Models} that {Generalize}},'' in
  \emph{\BIBforeignlanguage{en}{International {Conference} on {Learning}
  {Representations} ({ICLR} 2021)}}, Mar. 2021, arXiv: 2006.08762. [Online].
  Available: \url{http://arxiv.org/abs/2006.08762}
\BIBentrySTDinterwordspacing

\end{thebibliography}

\end{document}